# Interpretable to Whom? A Role-based Model for Analyzing Interpretable Machine Learning Systems

Richard Tomsett [1]  Dave Braines [1,2]  Dan Harborne [2]  Alun Preece [2]  Supriyo Chakraborty [3]

## Abstract

Several researchers have argued that a machine learning system's interpretability should be defined in relation to a specific agent or task: we should not ask if the system is interpretable, but *to whom* is it interpretable. We describe a model intended to help answer this question, by identifying different roles that agents can fulfill in relation to the machine learning system. We illustrate the use of our model in a variety of scenarios, exploring how an agent's role influences its goals, and the implications for defining interpretability. Finally, we make suggestions for how our model could be useful to interpretability researchers, system developers, and regulatory bodies auditing machine learning systems.

## 1. Introduction

"Interpretability" is a current hot topic in machine learning research. The increasing complexity of modern machine learning systems and the models they use, a wider adoption of machine learning in a variety of real-world systems, and new laws defining citizens' rights in relation to data processing systems (Doshi-Velez et al., 2017) all contribute to this heightened interest. Many groups have developed techniques intended to improve machine learning systems' interpretability (Zhang & Zhu, 2018; Guidotti et al., 2018; Chakraborty et al., 2017; Rudin, 2014), though the definitions of and motivations for interpretability (if specified) vary wildly between methods (Lipton, 2016).

Several researchers are trying to address this problem by formalizing the study of machine learning interpretability. Freitas made an early contribution, analyzing five classification models and discussing possible measures for interpretability (Freitas, 2014). Lipton notes that a model requires better interpretability when its predictions, and the metrics calculated on those predictions, are insufficient for characterizing it. He provides a taxonomy for categorizing interpretability methods with different properties (Lipton, 2016). Doshi-Velez and Kim expand on this motivation: "the need for interpretability stems from an incompleteness in the problem formalization, creating a fundamental barrier to optimization and evaluation" (Doshi-Velez & Kim, 2017), and provide a taxonomy for evaluating model interpretability. Miller reviews approaches to interpretability developed in philosophy and social science, discussing how artificial intelligence interpretability researchers could build on this existing literature (Miller, 2017). Poursabzi-Sangdeh et al. performed pre-registered experiments that measured the effect of different interpretability methods on user trust, ability to simulate models, and ability to detect mistakes (Poursabzi-Sangdeh et al., 2018). Dhurandhar et al. developed, to our knowledge, the first formal, quantifiable definition of interpretability (Dhurandhar et al., 2017b;a), which proposes measuring the interpretability of a procedure in relation to the performance of a *target model* (which could be a human or non-human agent) on a specific task. Most recently, Ras et al. analyzed what can be explained in relation to expert and lay-users of deep neural networks, providing a taxonomy for explanation methods (Ras et al., 2018).

In this paper, we explore *to whom* a machine learning system might be interpretable. While others previously identified that interpretability should be considered with reference to a specific user or user group (Kirsch, 2017), we develop this insight into a model for analyzing machine learning systems and the agents they interact with or affect. Agents have different beliefs and goals depending on their roles in relation to the machine learning system. Our model can be used to identify an agent's relationship to the system, and thus guide the analysis of what their relevant beliefs and goals might be for specifying suitable measures of interpretability for that agent.

[1] IBM Research, Hursley, Hampshire, UK [2] Crime and Security Research Institute, Cardiff University, Cardiff, UK [3] IBM Research, Yorktown Heights, New York, USA. Correspondence to: Richard Tomsett <rtomsett@uk.ibm.com>.

## 1.1. Definitions

Before outlining the model, we define how we intend to use some relevant terminology. Our model is built around a *machine learning system*, by which we mean a system that includes one or more machine learning models, the data used to train the model(s), any interface used to interact with the model(s), and any relevant documentation. The system (when unambiguous, we use just "system" to mean "machine learning system") could be monolithic, or comprised of several different services owned by different entities, situated in different locations, and trained on data from many sources.

The system is situated in a *machine learning ecosystem* (or just "ecosystem"), which includes the system and the agents that have interactions with, or are affected by, this system. An ecosystem always contains just one machine learning system and one or more agents (in the real-world, ecosystems will often overlap).

We define what we mean by an explanation and an interpretation:

- *Explanation*: the information provided by a system to outline the cause and reason for a decision or output for a performed task – a "post-hoc explanation" in Lipton's taxonomy (Lipton, 2016).

- *Interpretation*: the understanding gained by an agent with regard to the cause for a system's decision when presented with an explanation.

Thus an agent forms an interpretation by examining one or more explanations from a system. Considering these definitions, we also define explainability and interpretability, as well as *transparency* as we use it in relation to interpretability:

- *Explainability*: the level to which a system can provide clarification for the cause of its decisions/outputs.

- *Transparency*: the level to which a system provides information about its internal workings or structure, and the data it has been trained with – this is similar to Lipton's definition of transparency (Lipton, 2016).

- *Interpretability*: the level to which an agent gains, and can make use of, both the information embedded within explanations given by the system and the information provided by the system's transparency level.

These definitions hint that explainability, transparency and interpretability should be quantifiable. Indeed, our informal definition of interpretability is compatible with Dhurandhar et al.'s quantitative $\delta$-interpretability framework (Dhurandhar et al., 2017b;a) mentioned earlier.

In the next section we develop a conceptual model of an ecosystem and identify the different roles that agents can play within it. We then use this model to identify and classify the different agent groups in several example scenarios, and discuss how each kind of agent will have a different view of system interpretability. We also consider the constraints that need to be applied for each role, e.g., to protect sensitive data, or to preserve the privacy of the system's internal algorithms.

## 2. Ecosystem model

We define six different roles for the agents in an ecosystem. These were identified through discussion of many different scenarios involving machine learning systems (a few of which we outline in the next section), and build on ideas from the literature (Weller, 2017; Doshi-Velez & Kim, 2017; Miller, 2017; Ras et al., 2018). The roles are not mutually exclusive: a single agent could occupy any combination of roles, though some combinations are more likely than others. Currently, we expect most of these roles to be fulfilled by humans. However, machines may increasingly occupy them in future, especially if artificial agents gain rights over their data.

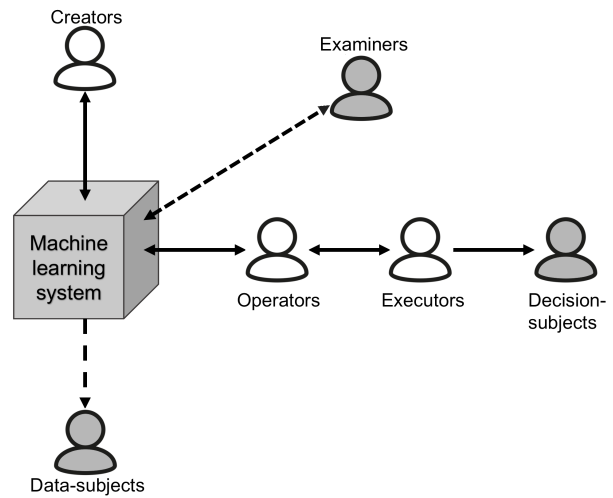

*Figure 1.* Illustration of a machine learning ecosystem. Direction of arrow indicates direction of interaction (e.g., data-subjects do not interact with the system, but the system has their data)

1. *Creators*: agents that create the machine learning system. Several teams of creators may work on different aspects of the same system e.g., architecture, design, implementation, training, documentation, deployment, and maintenance. Ecosystems always contain creators. When necessary, we further make the distinction between creator-owners and creator-implementers:

9

- *Owners*: the agent(s) or organization(s) that own the intellectual property in the machine learning system.
- *Implementers*: the agent(s) that directly implement the system, usually employees of or contractors for the owners

2. *Operators*: agents that interact directly with the machine learning system. Operators provide the system with inputs, and directly receive the system's outputs. In some cases they may be able to interact directly with the system's creators. Ecosystems always contain operators.

3. *Executors*: agents who make decisions that are informed by the machine learning system. Executors receive information from operators. Ecosystems always contain executors.

4. *Decision-subjects*: agents who are affected by decision(s) made by the executor(s). Ecosystems usually contain decision-subjects.

5. *Data-subjects*: agents whose personal data has been used to train the machine learning system. Ecosystems only contain data-subjects if the machine learning system has been trained on personal data.

6. *Examiners*: agents auditing or investigating the machine learning system. Depending on the system, they may interact with one or more of the other roles and the machine learning system itself. Ecosystems only contain examiners when the system is being audited/inspected.

We developed this categorization to help inform our design of the requirements for interpretability in different scenarios. The role an agent fulfills will impact its goals within a particular scenario, and thus its conception of the machine learning system's interpretability. We illustrate this by outlining some example scenarios.

## 3. Example scenarios

*Scenario 1: web advertising*

Many web-sites offer paid advertising spaces that operate via auction. When a user visits such a site, advertisers bid for ad space based on their valuation of that user seeing their advertisement. They estimate this with models that take as inputs user-data sent by the host web-site. The highest bidder's advert is displayed on the web-site. The ecosystem in this scenario contains a machine learning system made up of models from many different advertisers, a host web-site that displays the advert depending on the bids from the system, and a user browsing the host web-site.

– Creators: The advertising company and its employees, any third-party development companies and their employees

– Operator: the host web-site

– Executor: the host web-site

– Decision-subject: the web-site user

– Data-subjects: any internet denizen whose data has been obtained by the advertising companies

– Examiners: relevant advertising standards body staff, "data commissioner" style authority staff (e.g., the UK's Information Commissioner's Office); usually, such authorities will only become examiners if a complaint or information request is made

*Scenario 2: route planning on a smartphone*

Most smartphones provide apps (often machine learning systems) for planning driving routes. The user enters a desired start and end location, and the app provides one or more possible routes for them to take. This ecosystem contains a user planning a route, and an app that generates possible routes.

– Creators: the navigation app company and its employees

– Operator: the app user

– Executor: the app user

– Decision-subject: the app user

– Data-subjects: any road users whose location data has been obtained and used by the navigation app company

– Examiners: "data commissioner" style authority staff; usually, such authorities will only become examiners if a complaint or information request is made

*Scenario 3: loan application*

When someone applies for a loan, the lender may use a machine learning system to determine the applicant's chance of defaulting, and adjust the the amount offered, rate of interest, and other terms accordingly (or simply refuse to lend). This ecosystem contains a lender, an applicant seeking a loan, and a machine learning system for assessing applicant risk.

– Creators: The lender and its employees if they also developed the system, or any third-party development companies and their employees

– Operators: the lender's (customer-facing) employees*

– Executor: the lender's (higher-ranking) employees*

– Decision-subject: the loan applicant

– Data-subjects: prior loan applicants, any agent whose data has been obtained by the lender (likely most financial service

users)

– Executors: financial regulation authority staff, financial ombudsman

*The operators will likely be customer-facing agents who interact directly with applicants. They will make a decision based on the machine learning system's output, but the business logic for this decision will have been decided on by more senior employees at the lender. The senior employees would be seen as the executors in this case, as the customer-facing agent simply communicates the decision to the applicant.

*Scenario 4: medical advice for clinicians*

Several machine learning systems have been developed to assist doctors with diagnosis and treatment planning. The doctor provides the system with a patient's data, and the system judges the most likely diagnosis, or recommends possible treatment options that the doctor can then discuss with the patient. This ecosystem contains a machine learning system for diagnosis and/or treatment recommendation, a doctor or team of medical professionals who operate the system, and a patient.

– Creators: the medical software company and its employees, any collaborating medical professionals and researchers

– Operators: medical professionals

– Executors: the patient, medical professionals†

– Decision-subject: the patient

– Data-subjects: other patients, researchers and study subjects (e.g., data loaded from publications)

– Examiners: professional medical authority staff e.g., the UK's General Medical Council

†The role of executor in this scenario is debatable, and produced some discussion among the authors. The doctor makes decisions on treatment options based on the system's advice, so can be considered an executor. They would also be held responsible for treatment decisions from the point of view of an Examiner. However, the patient ultimately decides on their treatment, so could also be considered the executor. Additionally, in some cases a patient may be unable to make such decisions — they may be minors, or adults not in a sound frame of mind. In these cases, the patient's representative(s) (e.g., parent/guardian, attorney-in-fact) may be the executor(s).

*Scenario 5: releasing defendants on bail*

Machine learning systems are used in some countries to predict the likelihood a defendant will be dangerous if released on bail. These predictions are used by the judge about whether to grant bail, and at what price. This ecosystem contains a judge, a defendant, and a machine learning system consulted by the judge.

– Creators: the legal software company and its employees

– Operators: the judge (or other court staff)

– Executor: the judge

– Decision-subject: the defendant

– Data-subjects: previous defendants

– Examiners: In the case of an appeal, the original decision may be scrutinized by, for example, the defendant's lawyers. In this scenario, these lawyers would become examiners.

*Scenario 6: go no-go order in a military operation*

Consider a scenario involving a military operation to kill or capture a target. A machine learning system may be employed to find and help recognize this target, and the order to engage will be informed based on the system's recognition. After the event, this order may be scrutinized at a tribunal. This scenario includes the target, front-line personnel, analysts, a mission commander, tribunal jurors, and a machine learning system that may be distributed across several coalition partners.

– Creators: employees of the various coalition partners' militaries, employees of any military contractors involved

– Operators: military analysts

– Executor: the mission commander

– Decision-subject: the target, or agent identified as the target

– Data-subjects: other individuals of interest, their known associates

– Examiners: tribunal jurors

# 4. Role-based interpretability

Having defined the roles and provided example scenarios, we consider what interpretability means for agents fulfilling each role by considering their goals, noting that these goals are not always aligned (Weller, 2017).

## 4.1. Creators

In our above scenarios, the creator-implementers are the architects, designers, engineers, and technical writers responsible for constructing the system, and the creator-owner is the organization that owns the intellectual property in the system. Creator-implementers generally work for a single owner, but may create systems as collaborative efforts between several owners. Creator-implementers may also include subject matter experts from their own or other organizations, who provide knowledge to help train the system

11

(e.g., medical professionals and researchers in scenario 4).

Creators will want to improve system performance, where performance is a handily vague catch-all term for a variety of metrics to optimize for. These metrics depend on the particular scenario, and might include predictive accuracy, computational or data efficiency, bias minimization (Cowgill & Tucker, 2017), and/or safety (Varshney & Alemzadeh, 2017; Amodei et al., 2016). Their interpretability goal will therefore be to improve their understanding of the system such that they can better optimize it for their preferred metrics. A good example of creators engaging in such research in the domain of autonomous vehicles is given in (Bojarski et al., 2017).

Explainability and transparency are both important for improving creator-interpretability.

### 4.2. Operators

If the operator is not also the executor, then the operator must pass on information to the executor to inform their decision. They want to make sure the data they input to, or question they ask of the system is the right one for them to provide useful information to the executor. They may present all the available information to the executor, or a summary of it, depending on the scenario and executor's characteristics, and may need to obtain further explanations from the system in response to queries from the executor. Existing explanation methods that might improve operator-interpretability include techniques to highlight relevant input features — e.g., LIME (Ribeiro et al., 2016), layer-wise relevance propagation (Montavon et al., 2017) — or those that generate text explanations for outputs (Hendricks et al., 2016; Park et al., 2016).

Explainability is important for operator-interpretability. Transparency will be important if the operator requires some understanding of the system's internals to make good queries.

### 4.3. Executors

Executors are responsible for decision making, so want to be sure that they make good decisions. What constitutes a "good" decision varies depending on the executor's desires and goals. In scenario 2, a good decision might be to follow the shortest recommended route, or it might be to follow a longer route if that route is more scenic and if the executor values this feature for the journey in question. In scenario 4, the best treatment option for a terminal patient will depend on the patient's preferences around life-extension and quality-of-life. In scenario 6, the order to engage will be assessed based on a range of factors including mission success, casualties, and any collateral damage. In each case, suitable explanations for the system's outputs could affect decision making. We have previously developed an example system for providing some of the different levels of explanation that might be required by an executor (Harborne et al., 2018).

Explainability is important for executor-interpretability. If the executor is not also the operator, it is unlikely that transparency will affect executor-interpretability significantly.

### 4.4. Decision-subjects

If the decision-subject is not also the executor, then they will want to know why an executor made a particular decision, either out of plain curiosity, or to be able to challenge or change that decision — see (Hirsch et al., 2017) for an exploration of this idea of *contestability*. In scenario 1, the web-site user may want to know why they were shown a particular advert for interest, or so that they can remove or hide their personal data to prevent targeted advertising (or even provide more data to receive better targeted advertising). In scenario 3, the loan applicant may want to know how change their behaviour to make the system give them a lower risk score. In scenario 5, the defendant may want to challenge a decision, e.g., on grounds of discrimination. In scenario 6, the enemy may want to know how to avoid detection by the system to prevent being shot.

The goals of decision-subjects can clearly clash with those of executors and creators; this goal mismatch was previously noted in (Weller, 2017). In scenario 1, the web-site owner wants to maximize advertising revenue, but users who hide their data are less valuable than users with a data-rich profile. In scenario 3, the lender may be concerned that an applicant could game the system if it is highly interpretable to them — Akyol et al. formally analyzed this kind of strategic behaviour (Akyol et al., 2016). We leave identifying the goal mismatch in scenario 6 as an exercise for the reader.

Explainability is important for executor-interpretability, while transparency may be important in some scenarios.

### 4.5. Data-subjects

Data-subjects may not even be aware that a system has been trained on their data. However, in many jurisdictions, data subjects will have certain rights over their personal data. Given the growing awareness of data collection and sharing between companies, data-subjects may become more likely to exercise these rights and ask organizations to delete some or all of their data, with the goal of increasing their privacy — see, e.g., (Chen et al., 2016) for a study analyzing Facebook's predictions about users before and after data deletion.

Data-subjects may have moral concerns about how their data is being used to make decisions about other people, so may want to understand how their data affects a machine learning system's outputs. If they request that their data be deleted, the system's behaviour may not change at all: for example, a neural network's weights will only update on re-training, but a k-nearest neighbour model uses the data directly so will be immediately affected. Data-subjects may want to know about these consequences before requesting data deletion, and indeed, may be given the right to request system re-training in the case that deletion does not immediately affect the system's behaviour.

As data-subjects do not directly see a system's outputs unless they are also an operator, only transparency affects data-subject-interpretability.

### 4.6. Examiners

We added the examiner role to model agents tasked with compliance/safety-testing, auditing, or forensically investigating a system. Safety-testing hopefully occurs before deployment, so testers examine possible future outputs, while auditors and forensic investigators examine past system outputs. The latter requires an interpretable system to store both its decisions and their explanations, or to be able to generate explanations for stored past decisions. If explanations are generated, these should be identical to the explanation given at the original decision time. The examiner may want to interact with the system using new data to explore its outputs and their explanations (e.g., creating repeated "what-if" scenarios to establish decision sensitivity), which, if done forensically, also requires the system to respond as it would have at the time of the original outputs. This constraint might be very costly in, e.g., reinforcement learning, where models are continuously updated in response to environmental feedback.

Some types of examiners may provide feedback to creators on how to improve the system. For example, they could suggest ways to retrain models using approaches designed to improve feature relevance (Ross et al., 2017), or to improve model fairness by regularization to ensure that certain sensitive attributes of the data-subjects are not used in model building, even if they are good discriminators (Kamishima et al., 2011).

Both explainability and transparency are important for examiner-interpretability.

## 5. Discussion

Our model is a first draft intended to stimulate discussion and suggestions for improvements, but we anticipate that it could be useful to interpretability researchers, system developers (creator-owners), and regulatory bodies. As noted previously, interpretability is a woolly concept with inconsistently applied terminology. We hope to contribute to the formalization of some of these terms by building on existing definitions, using them consistently, and describing how they relate to the roles in our ecosystem model. Such a formalization is crucial for progressing machine learning interpretability research as a rigorous science (Doshi-Velez & Kim, 2017), allowing appropriate, quantitative comparisons to be made between similar methods under well-defined circumstances.

System creator-owners will ultimately decide on the extent to which they make their systems explainable and transparent to different roles. Our model could help creator-owners identify the interpretability needs of different agents in the ecosystem, allowing for a more systematic analysis and indicating where to focus their research and development efforts depending on their own goals for the ecosystem. The division of roles also helps to identify agents with conflicting goals, so that the owners can plan how best to manage this. They may develop a privacy model for the system around our ecosystem model, specifying different levels of access to various kinds of explanation depending on an agent's role (e.g., not allowing "explanations by analogy" (Lipton, 2016) that would reveal personal data to particular roles, or limiting transparency to non-creators to protect intellectual property).

Finally, regulatory bodies, filling the role of examiners, may use our model to aid in their auditing or forensic investigations of ecosystems. They may focus on decision-subjects and data-subjects, ensuring that the system is compliant with their rights — such as personal data privacy or the "right to an explanation" under GDPR (Goodman & Flaxman, 2016) — or examine why executors made specific decisions and the role of the machine learning system in influencing those decisions (e.g., identifying bias).

## Acknowledgment

This research was sponsored by the U.S. Army Research Laboratory and the UK Ministry of Defence under Agreement Number W911NF–16–3–0001. The views and conclusions contained in this document are those of the authors and should not be interpreted as representing the official policies, either expressed or implied, of the U.S. Army Research Laboratory, the U.S. Government, the UK Ministry of Defence or the UK Government. The U.S. and UK Governments are authorized to reproduce and distribute reprints for Government purposes notwithstanding any copy-right notation hereon.

13